\documentclass[runningheads]{llncs}
\usepackage[T1]{fontenc}
\usepackage{graphicx}
\usepackage{acronym}
\acrodef{ANN}{Artificial Neural Network}
\acrodef{CCA}{Canonical Correlation Analysis}
\acrodef{CNN}{Convolutional Neural Network}
\acrodef{DL}{Deep Learning}
\acrodef{DNN}{Deep Neural Network}
\acrodef{ML}{Machine Learning}
\acrodef{MLP}{Multi-Layer Perceptron}
\acrodef{NAP}{Neuron Activation Profile}
\acrodef{PCA}{Principal Component Analysis}
\acrodef{XAI}{Explainable Artificial Intelligence}

\usepackage{url}
\usepackage{subcaption}
\usepackage{todonotes}
\usepackage{amsmath}
\usepackage{amsfonts}
\usepackage{hyperref}
\usepackage[nameinlink]{cleveref}
\usepackage{natbib}

\Crefname{pluralfig}{Figures}{Figures}

\usepackage{booktabs}
\begin{document}
\title{Assessing Intersectional Bias in Representations of Pre-Trained Image Recognition Models}
\titlerunning{Intersectional Bias in Pre-Trained Image Recognition Models}
%
\author{Valerie Krug \and
Sebastian Stober}
\authorrunning{V. Krug and S. Stober}
%
\institute{Artificial Intelligence Lab, Otto von Guericke University Magdeburg, Germany\\
\email{\{valerie.krug,stober\}@ovgu.de}}
\maketitle              
\begin{abstract}
Deep Learning models have achieved remarkable success. 
Training them is often accelerated by building on top of pre-trained models which poses the risk of perpetuating encoded biases. 
Here, we investigate biases in the representations of commonly used ImageNet classifiers for facial images while considering intersections of sensitive variables age, race and gender.
To assess the biases, we use linear classifier probes and visualize activations as topographic maps. 
We find that representations in ImageNet classifiers particularly allow differentiation between ages. 
Less strongly pronounced, the models appear to associate certain ethnicities and distinguish genders in middle-aged groups.

\end{abstract}

\section{Introduction}

\ac{DL} has proven to be a highly effective tool across numerous fields of application \cite{dosovitskiy2020image,kim2022squeezeformer}. 
One key factor behind its success is the ability of \ac{DL} models to recognize patterns and relationships within data. 
However, this strength can also introduce a significant risk: the information these models learn can include unwanted biases from the data. 
Biases embedded in training data can reinforce stereotypes, leading to predictions that are inaccurate or unfair, particularly toward marginalized groups \cite{bolukbasi2016man}.
These biases can impact not only the model's outputs but also the internal representations it develops for certain groups. 
To address this issue, researchers have developed methods aimed at detecting and reducing biases in \ac{DL} models \cite{hort2023bias,ahn-oh-2021-mitigating}. 
Many of these approaches focus on specific aspects such as race, gender, or age -- referred to as sensitive variables. 
However, these variables are often observed independently which overlooks the interactions between different factors, commonly known as intersectionality.
For instance, even if a model shows no bias against women and people of color as general categories, it might still exhibit bias against women of color, an issue that could remain undetected without examining intersecting variables. 
This highlights the importance of considering the combined influence of multiple sensitive variables when addressing bias in \ac{DL} systems.

The development of \ac{DL} models, especially in image recognition, frequently relies on a technique called transfer learning \cite{torrey2010transfer}, where pre-trained models are used as basis for an own learning task. 
Transfer learning facilitates the development of new models by reducing the demands for large datasets and computational resources. 
However, biases embedded in the data used to train the original models can easily be propagated \cite{salman2022does}. 
If these pre-trained models contain biases, they may be inherited and potentially amplified in the models that build upon them. 
To ensure fairness while utilizing the benefits of transfer learning, it is essential to evaluate pre-trained models thoroughly. 
Identifying the biases they encode is a critical step in understanding and potentially mitigating inequalities that arise from applying transfer learning with particular pre-trained models.

In previous research \cite{krug2023topomapsbias,krug2024thesis}, we have investigated biases in pre-trained image recognition models with a focus on their representations. 
We examined bias in the neuron activations of the VGG16 model for sensitive attributes ``race'', ``age'', and ``gender'' with a visualization-based approach.
However, more advanced image recognition models than VGG16 are available and their studies considered the sensitive variables independently.
Here, we aim to expand on these experiments. 
Firstly, we include a broader range of \ac{CNN} models to identify which biases are independent of the architecture and hence derive from the data.
Secondly, given the importance of accounting for intersectionality, we focus on the intersections of sensitive attributes to provide a more comprehensive analysis. 
To this end, we use a visualization-based analysis that allows the visual inspection of representations, and introduce novel qualitative and quantitative bias measures based on linear classifier probes.
The Code is publicly available\footnote{\url{https://github.com/valeriekrug/iNNtrospect/tree/intersectional-bias}}.

\section{Background}
\subsection{Deep Neural Networks}
\ac{DL} is a type of \ac{ML} and employs \acp{ANN}, specifically those with multiple layers, commonly known as \acp{DNN} \cite{goodfellow2016deep}. 
In \acp{DNN}, artificial \textit{neurons} play a crucial role in processing information.
Each neuron computes a weighted sum of its connected inputs and applies a non-linear activation function to the result. 
The outcome of this function is referred to as the \textit{activation} of the corresponding neuron. 
The weights between the neurons, used to compute the activation, are the main trainable parameters of a \ac{DNN}.
Neurons are organized in hidden \textit{layers}, where neurons within the same layer are not interconnected.

In this work, we focus on a particular type of \ac{ANN} architecture, \acfp{CNN}.
These models can detect patterns irrespective of their position in the input. 
They learn small sets of weights, called \textit{filters}, applied across the entire input, allowing them to efficiently detect patterns at any position. 
The output of a single convolutional filter applied to the entire input is referred to as a \textit{feature map}. 
Individual neurons are the positions of feature maps, and due to the convolutional architecture, not every pair of neurons in subsequent layers is connected. 
Each layer in a \ac{CNN} can learn multiple filters, each generating a corresponding feature map. 
\acp{CNN} are particularly suitable for handling data with spatial or temporal relationships, such as images or sequential data.

\subsection{Transfer Learning}
Training \acp{DNN} from scratch presents challenges, especially when computational resources are scarce or data availability is limited. 
In such low-resource scenarios, a prevalent strategy is to employ transfer learning, where an existing pre-trained model is adapted for a different task. 
Transfer learning is based on the concept that similar data share fundamental features, which only are arranged differently in specific contexts.
A notable application of transfer learning is in image recognition, where various models pre-trained on large-scale datasets are openly available. 
Such models serve as feature extractors, excluding their top classification layer(s). 
The extracted features can then be used as input to a smaller network which is trained to perform some other image processing task. 
This approach reduces the necessity for extensive training on large data sets because only few (new) layers require adaptation.

Transfer learning is a powerful tool for using information from pre-trained models to solve new tasks.
However, it poses the risk that the biases present in the dataset that is used to train a model may be reproduced.
For example, if a pre-trained vision model was trained on data that predominantly include images of certain demographic groups or cultural contexts, it may inadvertently learn biases associated with those groups. 
When this model is adapted for a new task, these biases can influence the new model's predictions, potentially leading to unfair or discriminatory outcomes.

\section{Related Work}

\subsection{Explainable Artificial Intelligence}
\ac{XAI} involves the examination or visualization of the internal workings of computational models. 
This is particularly relevant in the context of \ac{DNN}, which often function as black boxes \cite{Yosinski2015}.
Efforts to elucidate the internal mechanisms of \ac{DL} models have experienced significant attention recently, leading to the development of various \ac{XAI} techniques.
They are frequently applied in computer vision, where the visual interpretation of image data is relatively easy \cite{Selvaraju2017}.

One common approach is to analyze the patterns learned by a model. 
For some models, this is possible by inspecting their weights \cite{Osindero2008}.
More complex models can be investigated by analyzing which inputs activate specific parts of the network. 
Such inputs can be examples from real datasets or synthetic inputs created through optimization \cite{Erhan2009,Mordvintsev2015}. 
Another popular strategy involves examining the output and determining the significance of input features for individual predictions, known as attribution techniques \cite{Selvaraju2017,Kindermans2018}.

Of particular relevance to this work are \ac{XAI} techniques that analyze \ac{DNN} activations and outputs across a multitude of inputs, often grouped by specific properties of interest such as class membership or, as in our work, by sensitive variables. 
This type of analysis provides a deeper understanding of how different groups or classes are represented within a \ac{DNN}.
For instance, linear classifiers applied to intermediate representations can quantify a layer's representative power for the prediction \cite{Alain2017}.
Moreover, these classifiers can be the basis for deriving vectors that represent user-defined concepts \cite{Kim2018}. 
Processes within \acp{DNN} can also be characterized by examining the similarities between activations across different inputs or groups of inputs. 
Such analyses can include techniques like \ac{PCA} \cite{Fiacco2019}, \ac{CCA} \cite{Morcos2018a}, dimensionality reduction \cite{carter2019activation,hohman2019s,park2021neurocartography}, or clustering of class-specific neuron activations \cite{Nagamine2015,krug2024thesis}.
For comparative analysis between groups, it is often beneficial to consider activation differences rather than the activations themselves \cite{krug2023visualizing}.

\subsection{Investigating Bias in DNNs}
\acp{DNN} have a tendency to reproduce or emphasize biases present in the training data. 
Various methods have been introduced to identify and address these biases. 
For instance, Sweeney \cite{sweeney2013discrimination} exposed racial discrimination in Google's online ad delivery, while Bolukbasi et al. \cite{bolukbasi2016man} worked on debiasing widely used word embeddings. 
More recently, researchers have explored biases in modern Transformer-based models like BERT \cite{devlin2019bert}, for example, focusing on gender bias \cite{li2021detecting} and ethnic bias \cite{ahn-oh-2021-mitigating}.
Discrimination in facial recognition algorithms was investigated by Buolamwini and Gebru \cite{pmlr-v81-buolamwini18a}, who introduced the Gender Shades dataset with a balanced distribution of gender and skin types. 
Similarly, the FairFace dataset \cite{karkkainenfairface} aims to provide a balance across gender, race, and age categories.

Many studies investigate bias in the output of algorithmic decision making systems \cite{pmlr-v81-buolamwini18a,karkkainenfairface}.
Different to that, our emphasis lies on the representations encoded in the individual layers of a \ac{DNN}, in order to reveal representational bias in pre-trained models while considering intersectionality.
Mitigating biases, however, is out of the scope of this work.

\section{Methods}

\subsection{Linear Classifier Probes}
\label{sec:probes}
Linear classifier probes \cite{Alain2017} aim to understand which information a \ac{DNN} has learned to encode in its layers.
This approach uses activations of a \ac{DNN} in a particular layer and is trained to predict categories from a concept of interest.
These target classes for the linear classifier can be different from those which the \ac{DNN} is trained on.
Better classification performance indicates that this concept is encoded in the representations of the investigated layer.

\subsection{Neuron Activation Profiles (NAPs)}
\label{sec:naps}
\acp{NAP} \cite{krug2024thesis} are a technique to characterize activations that are common to a certain group of inputs through averaging.
A group $G$ can be any set of input examples.
With $A_l()$, we denote the activations in a particular layer $l$, where $A_0()$ are the activations in the input layer, which is, the input examples themselves.
First, in a layer $l$ and for a group $G$, we compute the average activation $\overline{A_l(G)}$ across all activations of the instances in $G$.
For better contrasting the activations between multiple groups $\mathbf{G}$, we normalize $\overline{A_l(G)}$ for each $G \in \mathbf{G}$ by subtracting the expected activation $\mathbb{E}(\overline{A_l})$ to obtain what is called a \ac{NAP} of group $G$ in layer $l$. 
\begin{equation*}
	\begin{gathered}
	\overline{A_l(G)} = \frac{1}{|G|} \sum_{a \in A_l(G)}a \quad\quad\quad\quad\quad 
	\mathbb{E}(\overline{A_l}) = \frac{1}{\displaystyle\sum_{G \in \mathbf{G}}|G|} \sum_{a \in A_l(G) \forall G \in \mathbf{G}}a \\
	NAP_l(G) = \overline{A_l(G)} - \mathbb{E}(\overline{A_l})
	\end{gathered}
\end{equation*}
For computational efficiency, $\mathbb{E}(\overline{A_l})$ can be computed as the group size-weighted average of \acp{NAP} of different groups.

\subsubsection{Visualization as Topographic Activation Maps}
Inspired by how brain activity is intuitively displayed as topographic maps, we layout \ac{CNN} feature maps to allow for a similar activation visualization.
Considering a layer, we first distribute feature maps in a 2D space such that feature maps of similar activations are close to each other with a UMAP projection \cite{mcinnes2020umap,mcinnes2018umap-software}. 
Then, we distribute the feature maps more evenly in this 2D space by treating them as particles that attract each other to close gaps and repel others to avoid two particles at the same position. 
For the set of particles $N$ with initial coordinates from the UMAP projection, we compute a force $f(i)$ for each particle $i \in N$
\begin{equation*} 
	\label{eq:local_force}
	\begin{gathered}
		f(i) = \frac{\displaystyle\sum_{j \in N\setminus i}{\left(attr(i,j) - rep(i,j)\right)}}{|N\setminus i|}  \\
		attr(i,j) = 1.5 \cdot \left(dist(i,j) + 1\right) ^ {-3}   \quad\quad\quad
		rep(i,j) = 15 \cdot e^{-(dist(i,j)/2)}
	\end{gathered}
\end{equation*}
where $dist()$ is the Euclidean distance of particle coordinates, and apply this force to adjust the position of the particles for 1000 iterations. 

Finally, we visualize \acp{NAP} according to the computed layout.
To this end, for each group, we assign each coordinate a color that reflects the average \ac{NAP} value of the corresponding feature map for the respective group.
Then, to avoid gaps in the visualization, we linearly interpolate the colors between the positions with a resolution of $100 \times 100$~px.
To define colors, we map all groups' \ac{NAP} values to a 0-symmetric continuous color scale from blue over white to red, such that equal colors represent the same value in every group.
Typically, the colors use the highest absolute value as the end of the color scale.
In cases where there are outliers of extremely high or low \ac{NAP} values, this range can be symmetrically decreased to improve the contrast of the visualization.

\section{Experimental Setup}

In this section, we describe our experiments for investigating biases in pre-trained ImageNet classifiers evaluated using the FairFace dataset.
Quantitatively, we apply linear classifier probes to investigate whether classifying the sensitive variables in intersection is possible from the representations of the model's intermediate layers.
Qualitatively, we inspect common classifier probe errors and investigate whether biases are visually observable from activations and their similarities.
The code is publicly available. 

\subsection{Pre-Trained Models}
We investigate several \ac{CNN} models pre-trained on ImageNet \cite{deng2009imagenet}, which is one of the most widely used datasets in the field of \ac{ML} and computer vision.
The data set covers a wide range of categories, ranging from animals and everyday objects to abstract concepts. 
It plays a major role in the success of deep learning models, particularly \acp{CNN}, as ImageNet serves as a key benchmark for training and assessing the performance of these models. 
Many pre-trained models on ImageNet have been made publicly available, allowing researchers and developers to build on top of these models.

In this work, we use the pre-trained models VGG16, ResNet50, and InceptionV3 obtained from the TensorFlow Keras applications module\footnote{\url{https://www.tensorflow.org/api_docs/python/tf/keras/applications/}}.
For each model, we consider the input and output layers.
Further, we select representative layers for our analysis that are outputs of major building blocks of the models, for example, after feature map concatenation or when reducing the dimensionality by applying a pooling operation.

VGG16 \cite{DBLP:journals/corr/SimonyanZ14a} is a feed-forward \ac{CNN} that is built of stacks of convolutional layers followed by max pooling layers to decrease feature map dimensionality.
Because max pooling limits the information that is passed forward in the network, we select the corresponding layers.
Further, we investigate the first fully-connected layer because it marks where the model changes from convolutional to linear operations.
In Keras, these layers are named ``block\{1--5\}\_pool'' and ``fc1''.

ResNet50 \cite{he2016deep} employs residual connections between particular layers.
This means, some layers' input is the sum of the previous layer and an earlier layer.
The ResNet50 model first applies convolution and max pooling, followed by 16 residual blocks of varying number of filters.
Finally, the output of the last block is flattened with global average pooling and fed to the classification layer.
As representative layers, we consider the outputs of each residual block and the initial and final pooling layers.
The names in Keras are ``pool1\_pool'', every ``conv*\_block*\_out''\footnote{* indicates a placeholder for every occurring index} and ``avg\_pool''.

InceptionV3 \cite{szegedy2016rethinking} uses so-called Inception modules to allow the \ac{DNN} to process the same input in different ways.
To this end, each module comprises multiple stacks of convolution and/or pooling layers of different filter sizes to the same input whose outputs are concatenated in the filter dimension.
The InceptionV3 architecture initially comprises a stack of convolutional and max pooling layers, followed by 11 varying Inception modules, of which the final output is global average pooled and fed to a classification layer.
As representative layers for our analysis, we use the output of the final max pooling because it is the input to the first Inception module (``max\_pooling2d\_1''), the outputs of all Inception modules (``mixed\{0--10\}'') and the result of the final global average pooling (``avg\_pool'').

The models have different input sizes.
In our experiments, we scale the inputs to a suitable size using bilinear interpolation.
As all images and input sizes are square, the aspect ratio is preserved without cropping or padding.

\subsection{Evaluation Data}
\label{sec:fairface}
FairFace \cite{karkkainenfairface} is a facial image dataset designed to investigate bias in existing algorithmic systems. 
Its primary motivation is to provide a balanced and diverse dataset that includes a representative distribution of faces across various ethnicities, age groups, and genders.
In this work, we use FairFace to investigate biases in different \ac{DL} models trained on the ImageNet dataset.
None of the labels provided in FairFace is a category predicted by these models and none of the examples has been used to train them.
Therefore, FairFace is suitable as an independent evaluation data set.
Further, we can use the large training split (86,744 examples) of the FairFace data set for evaluating the models. 

\begin{figure}[!htb]
	\centering
	\begin{subfigure}{\linewidth}
		\includegraphics[width=\linewidth]{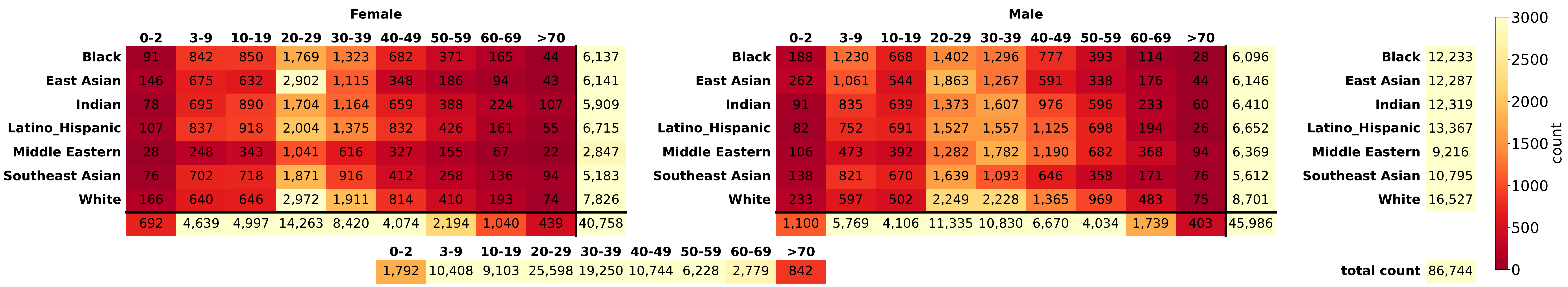}
		\caption{Entire training set. Heatmap colors are scaled between 0 and the largest intersectional group: ``White, 20--29, Female''.}
		\label{fig:var_freqs_total}
	\end{subfigure}
	\begin{subfigure}{\linewidth}
		\includegraphics[width=\linewidth]{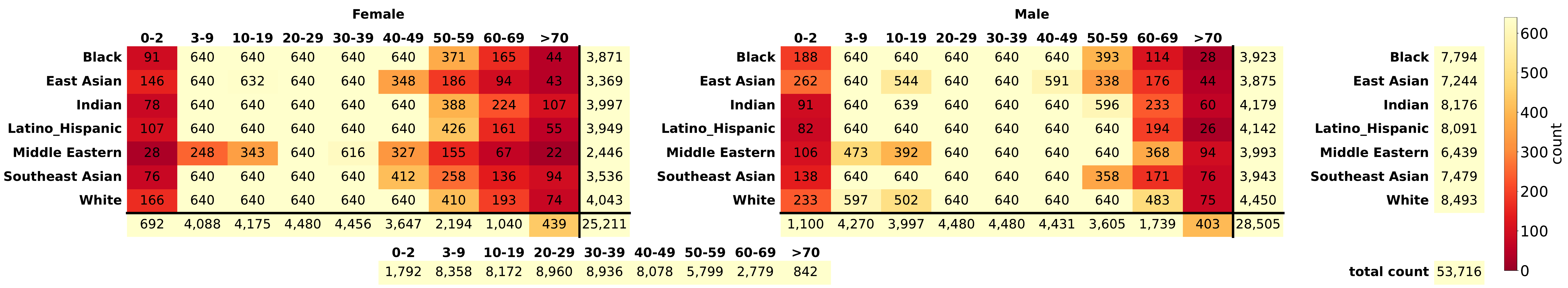}
		\caption{Capped at 640 examples per subgroup. Heatmap colors are scaled between 0 and 640.}
		\label{fig:var_freqs_capped}
	\end{subfigure}
	
	\caption{FairFace data set example counts. 
		Main tables: Image counts of the intersectional groups. 
		Tables' bottom rows: Number of examples of the respective ``age'' and ``gender'' (column sums). 
		Tables' last columns: Number of examples of the respective ``race'' and ``gender'' (row sums).
		Bottom right corners of each table: Number of images labeled ``Female'' or ``Male''.
		Single bottom row: Number of images in the ``age'' groups.
		Single right column: Image count for the ``race'' categories.
		Bottom right cell: Total number of images.
		The number of ``age'' and ``race'' combination groups, summed over the two gender categories is omitted to improve clarity of the figure.}
	\label{fig:var_freqs}
\end{figure}

For each image, FairFace provides labels for the sensitive variables $\textbf{V}=\{$``race'', ``age'', ``gender''$\}$  from 7, 9 and 2 categories, respectively.
We investigate the combination of the variables to account for intersectional biases.
Considering all combinations of ``race'', ``age'' and ``gender'', we obtain 126 groups $G \in \mathbf{G}$.
An overview of the number of examples in each of these intersectional categories is shown in \Cref{fig:var_freqs_total}.
Each group represents a combination of categories, denoted as ``$<$race$>$, $<$age$>$ ,$<$gender$>$'', for example, ``White, 20--29, Female'', and comprises the set of instances that belong to the corresponding sensitive variable combination.

Although FairFace aims to provide a balanced data set with respect to the three variables, we observe that the training set is not balanced regarding the intersection of categories.
The group sizes range from only 22 images for ``Middle Eastern, $>$70, Female'' to 2,972 images of ``White, 20--29, Female'' people.
Also, the age categories are imbalanced, ranging from only 842 images in the ``$>$70'' category to 25,598 in the age range of ``20--29''.
To decrease this imbalance, we limit the group size to a maximum of 640 examples in our experiments, leading to the distribution shown in \Cref{fig:var_freqs_capped}.
For groups that exceed this number, we select random examples from the group.

\subsection{Linear Classifier Probes}
\label{sec:experiment_probes}
With linear classifier probes  (see \Cref{sec:probes}), we investigate whether the intersectional groups $\mathbf{G}$ can be linearly distinguished, given their activations in a particular layer $A_l()$.
If the layer encodes information about these groups, a linear classifier could reach a high accuracy and would indicate that the representation is biased.

\subsubsection{Obtain Training and Validation Data}
We obtain data to train and evaluate the linear classifiers using a balanced set of examples from all 126 groups.
From each group, we select a batch of up to 128 random examples.
We split each batch randomly into 90\% for training and 10\% for validation.
Using the activations with their full resolution can lead to huge numbers of parameters of the linear classifier (e.g., 101,154,816 parameters for ResNet50 layer 2), which cannot be properly fitted with the small amount of data.
Assuming that there is similar information in spatial neighborhoods of activations in feature maps, we reduce their resolution in every channel to a maximum of 8$\times$8.
We investigate subsampling, i.e., taking each $n$-th value in both spatial dimensions, and average-pooling to reduce the resolution.

\subsubsection{Train and Evaluate a Linear Classifier}
We implement each linear classifier as a model that first flattens the activations and applies dropout with a rate of 0.25.
The classification is performed with a fully connected layer with 126 output neurons and L1~(1e-5) and L2~(1e-4) regularization.
We fit the model with categorical crossentropy and Adam optimizer with default parameters for 50 epochs.
To measure performance, we record training and validation accuracy over training time.
Further, we investigate common errors made by the linear classifier.

\subsubsection{Interpretation Instructions}
\label{sec:interpret_results_probes}
\begin{figure}[!htb]
	\centering
	\begin{minipage}{.5\textwidth}
		\centering
		\includegraphics[height=9em]{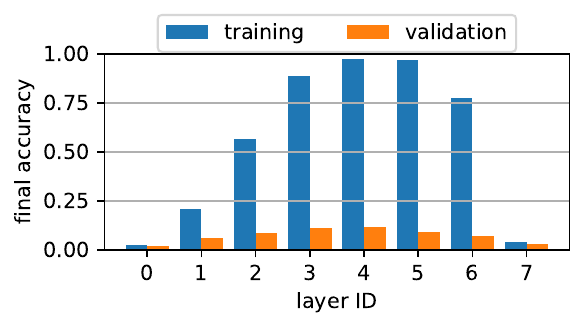}
		\caption{Exemplary accuracies along layers.}
		\label{fig:probe_tutorial_overview}
	\end{minipage}%
	\begin{minipage}{.5\textwidth}
		\centering
		\includegraphics[height=9em]{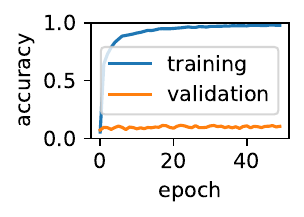}
		\caption{Exemplary learning curves in one layer.}
		\label{fig:probe_tutorial_layer}
	\end{minipage}
\end{figure}

We report the training and test accuracy at the end of training in a plot as shown in \Cref{fig:probe_tutorial_overview}.
Generally, layers with a higher accuracy indicate a bias in the respective layer.
In the example, we observe less bias in early and late layers than in the middle layers.
None of the classifiers generalizes well to unseen instances, but validation accuracy shows the same trend.
Layers 3 and 4 show relatively high validation accuracy and training performance is best for layers 4 and 5.
Hence, Layer 4 can be considered to encode biases most strongly comparing the layers.
The large gap between training and validation accuracy might be related to overfitting at a certain point in training.
To investigate this, we observe the performance over training time in learning curves as shown in \Cref{fig:probe_tutorial_layer}.
In the provided example, the classifier does not experience a drop of validation accuracy at any point.

\vspace{-0.5cm}
\begin{table}[!htb]
	\centering
	\def\arraystretch{1.1}
	\begin{tabular}{cccllcccllr}
		\toprule
		\multicolumn{3}{c}{label} &&& \multicolumn{3}{c}{prediction} &&& error rate for  \\
		race & age & gender &&& race & age & gender &&& target label \\
		\midrule
		Southeast Asian & \textbf{30--39} & Male &&& Southeast Asian & \textbf{40--49} & Male &&& 12.17\% \\
		Southeast Asian & \textbf{20--29} & Male &&& Southeast Asian & \textbf{40--49} & Male &&& 11.30\% \\
		\bottomrule
	\end{tabular}
	\vspace{0.2cm}
	\caption{Exemplary frequent errors of a linear classifier probe trained on one layer.}
	\label{tbl:probe_tutorial}
\end{table}
\vspace{-0.5cm}

Finally, in a layer of interest, we consider the most frequent misclassifications made by a linear classifier and display them as shown in \Cref{tbl:probe_tutorial}.
We further highlight which variables differ between the correct group and the prediction in bold print.
In the example, we see that the most likely errors are made by predicting the wrong ``age'' group, while ``race'' and ``gender'' are correct. 
Particularly, the classifier appears to wrongly classify groups as age ``40--49''. 

\subsection{Neuron Activation Profile Analysis}
\label{sec:experiment_NAPs}
\acp{NAP} follow the idea of aggregating activations over groups and then comparing their similarities (see \Cref{sec:naps}).
Obtaining such \ac{NAP} as a single characteristic activation pattern for every group facilitates visualization of the activations, which we specifically implement as topographic activation maps.
This analysis provides a comparative overview of activations for the intersectional groups and allows to visually identify potential biases.

\subsubsection{Compute and Visualize Neuron Activation Profiles}
We compute \acp{NAP} for the 126 intersectional groups, including all examples capped at 640 as shown in \Cref{fig:var_freqs_capped}.
In addition, we consider unions of groups that share one or two sensitive variables.
However, we normalize every $\overline{A_l(G)}$ with the same $\mathbb{E}(\overline{A_l})$ obtained from the 126 intersectional groups because they experience less intra-group variation and lead to more realistic expected activations.
We further visualize the \acp{NAP} as topographic maps.
To this end, we compute the neuron layout based on the intersectional groups' \acp{NAP}.
To increase color contrast, we use either the 0.5 or 99.5 percentile over all \ac{NAP} values, depending on which has the higher absolute value, to define the ends of the color scale.
Finally, we arrange the topographic maps of all groups in a tabular grid equivalent to the frequency table in \Cref{fig:var_freqs_capped}.

\subsubsection{Interpretation Instructions}

\label{sec:interpret_results_NAPs}
Interpreting topographic activation maps is based on comparing activations between groups.
If two groups have a similar activation pattern, they are expected to be difficult to distinguish for a model.

\begin{figure}[!b]
	\centering
	
	\begin{subfigure}{0.45\linewidth}
		\centering
		\includegraphics[height=6.5em, trim=432 5 650 420, clip]{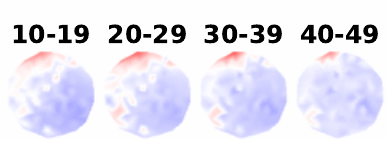}
		\caption{Exemplary similar ``age'' categories.}
		\label{fig:topomap_tutorial_allsame}
	\end{subfigure}
	\hspace{0.05\linewidth}
	\begin{subfigure}{0.45\linewidth}
		\centering
		\includegraphics[height=6.5em, trim=335 5 785 420, clip]{figures/VGG16/plots/topomap/topomap_overview_layer002.pdf}
		\includegraphics[height=6.5em, trim=620 5 555 420, clip]{figures/VGG16/plots/topomap/topomap_overview_layer002.pdf}
		\caption{Exemplary dissimilar ``age'' categories.}
		\label{fig:topomap_tutorial_different}
	\end{subfigure}
	
	\begin{subfigure}{1\linewidth}
		\centering
		\includegraphics[height=6.5em, trim=335 5 505 420, clip]{figures/VGG16/plots/topomap/topomap_overview_layer002.pdf}
		\caption{All ``age'' categories as topographic activation maps.}
		\label{fig:topomap_tutorial_somesame}
	\end{subfigure}
	
	 \caption{Exemplary sets of topographic activation maps from the \ac{NAP} analysis.}
\end{figure}

Firstly, if each category leads to a similar activation pattern, it is difficult to distinguish them and therefore, it indicates small bias for this variable.
An exemplary set of such topographic maps is shown in \Cref{fig:topomap_tutorial_allsame}.

Notably, such similarity can also indicate a bias when observing a different set of groups.
To illustrate this effect, we consider all ``age'' categories as shown in \Cref{fig:topomap_tutorial_somesame}.
Still, categories from 10--49 are similar to each other, but they differ from other categories.
Having a subset of groups that is similar while others are different indicates that it is likely for the model to confuse the similar groups.
This can be interpreted as a bias from merging categories that are independent.

Finally, activation patterns can vary between all groups, as illustrated in \Cref{fig:topomap_tutorial_different}.
On the one hand, this indicates bias because it is easy to distinguish the groups.
However, it reduces the likelihood of transferring decisions from one group to another.
Hence, there is at least some fairness because the activations allow to account for every individual group.

In summary, differences in activation patterns generally indicate biases.
Whether this bias is detrimental depends on the task at hand.
For a sensitive variable, either pairwise dissimilar or pairwise similar activation patterns can be desirable.
Pairwise similarity is beneficial if the aim is to not be able to distinguish the groups, like granting a loan independent of race or gender.
Dissimilarity between every pair of groups is a useful basis for tasks that need group-specific decisions without favoring one group over a set of others, like suggesting personalized medical treatments.

\section{Results and Discussion}

\subsection{VGG16}
\subsubsection{Layer Overview}

\captionsetup[subfigure]{format=plain}
\begin{figure}[!b]
	\centering
	\begin{subfigure}{0.4\linewidth}
		\includegraphics[width=\linewidth]{figures/classifier_probes_pool/accuracy_VGG16_overview.pdf}
		\caption{Accuracies using average pooling}
	\end{subfigure}
	\begin{subfigure}{0.4\linewidth}
		\includegraphics[width=\linewidth]{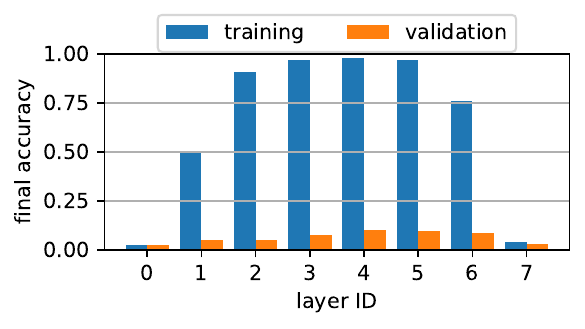}
		\caption{Accuracies using subsampling}
	\end{subfigure}
	
	\begin{subfigure}{0.24\linewidth}
		\includegraphics[width=0.9\linewidth]{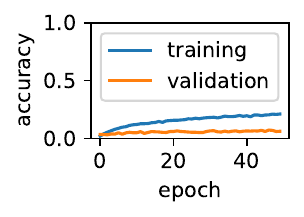}
		\caption[format=hang]{Avg. pooling, layer 1}
	\end{subfigure}
	\begin{subfigure}{0.24\linewidth}
		\includegraphics[width=0.9\linewidth]{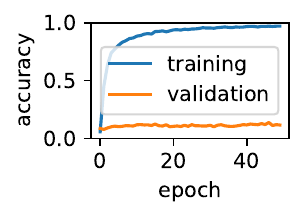}
		\caption{Avg. pooling, layer 4}
	\end{subfigure}
	\begin{subfigure}{0.24\linewidth}
		\includegraphics[width=0.9\linewidth]{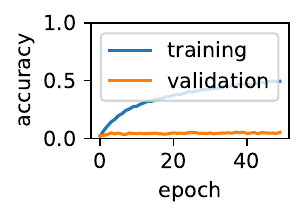}
		\caption{Subsampling, layer 1}
	\end{subfigure}
	\begin{subfigure}{0.24\linewidth}
		\includegraphics[width=0.9\linewidth]{figures/classifier_probes_subsample/accuracy_VGG16_layer004.pdf}
		\caption{Subsampling, layer 4}
	\end{subfigure}
	\caption{Linear classifier probe accuracies along the VGG16 layers and selected learning curves using different downsampling approaches.}
	\label{fig:VGG16_probe_overview}
	\label[pluralfig]{figs:VGG16_probe_overview}
\end{figure}

Training and validation accuracies at the end of the training are shown in \Cref{fig:VGG16_probe_overview}, using (a) average pooling and (b) subsampling for downsampling feature maps.
We observe an increase in accuracy from layer 0 to 4 and again a decrease until layer 7.
Particularly, the classifier cannot distinguish the groups in the input and output layers. 
Training and validation follow the same trend along the layers but the validation performance is significantly worse.
This indicates that the bias exists but is not strong enough for generalized linear separation.
From the learning curves (examples in \Cref{figs:VGG16_probe_overview}c--f), we find that there are no local maxima of the validation accuracy with better generalization.

Comparing average pooling and subsampling, there is clearly lower training and slightly higher validation accuracy for earlier layers using average pooling while there is similar accuracy in deeper layers.
This indicates that average pooling removes more information than subsampling in lower layers.
Considering that the feature maps are larger in lower layers, we suspect that this holds for every layer but becomes negligible for smaller feature maps.

\subsubsection{Classification Errors}
For layer 4 of VGG16, which has the highest bias according to the classifier accuracy, \Cref{tbl:errors_VGG16_layer004} shows the most common errors made by the linear classifier.
We observe two kinds of common errors.
The most prominent ones are wrong classifications as ``Indian, $>$70, Male'', which happen for groups of either slightly younger age (``60--69'') or other ethnicities (``Latino\_Hispanic'', ``Southeast Asian'' or ``White'').
Secondly, several groups are commonly misclassified as age group ``50--59'' with no visible pattern in the true group labels.
Gender prediction is mostly correct.
For this layer, we therefore suspect that the high bias particularly lies in the age and race variables.

\begin{table}[!htb]
	\def\arraystretch{1.1}
	\centering
		\begin{tabular}{cccllcccllr}
			\toprule
			\multicolumn{3}{c}{label} &&& \multicolumn{3}{c}{prediction} &&& error rate for  \\
			race & age & gender &&& race & age & gender &&& target label \\
			\midrule
			Indian & \textbf{60--69} & Male &&& Indian & \textbf{$>$70} & Male &&& 15.65\% \\
			\textbf{Latino\_Hispanic} & $>$70 & Male &&& \textbf{Indian} & $>$70 & Male &&& 13.04\% \\
			Indian & \textbf{60--69} & \textbf{Female} &&& Indian & \textbf{$>$70} & \textbf{Male} &&& 6.96\% \\
			\textbf{White} & $>$70 & Male &&& \textbf{Indian} & $>$70 & Male &&& 5.97\% \\
			\textbf{White} & \textbf{60--69} & Male &&& \textbf{Indian} & \textbf{$>$70} & Male &&& 5.22\% \\
			\textbf{East Asian} & \textbf{40--49} & Female &&& \textbf{Middle Eastern} & \textbf{50--59} & Female &&& 5.22\% \\
			\textbf{Latino\_Hispanic} & \textbf{$>$70} & Male &&& \textbf{White} & \textbf{50--59} & Male &&& 4.35\% \\
			\textbf{Latino\_Hispanic} & \textbf{$>$70} & Male &&& \textbf{Southeast Asian} & \textbf{50--59} & Male &&& 4.35\% \\
			Indian & \textbf{60--69} & Female &&& Indian & \textbf{50--59} & Female &&& 4.35\% \\
			Indian & \textbf{50--59} & Male &&& Indian & \textbf{$>$70} & Male &&& 3.48\% \\
			\bottomrule
		\end{tabular}
\vspace{0.2cm}
	\caption{Frequent training set errors of a linear classifier probe trained on layer 4 of VGG16. Categories in bold print indicate differences between annotation and predicted group.}
	\label{tbl:errors_VGG16_layer004}
\end{table}

\vspace{-1cm}

\subsubsection{Activation Visualization} 
\begin{figure}[!htb]
	\centering
	\includegraphics[width=\linewidth]{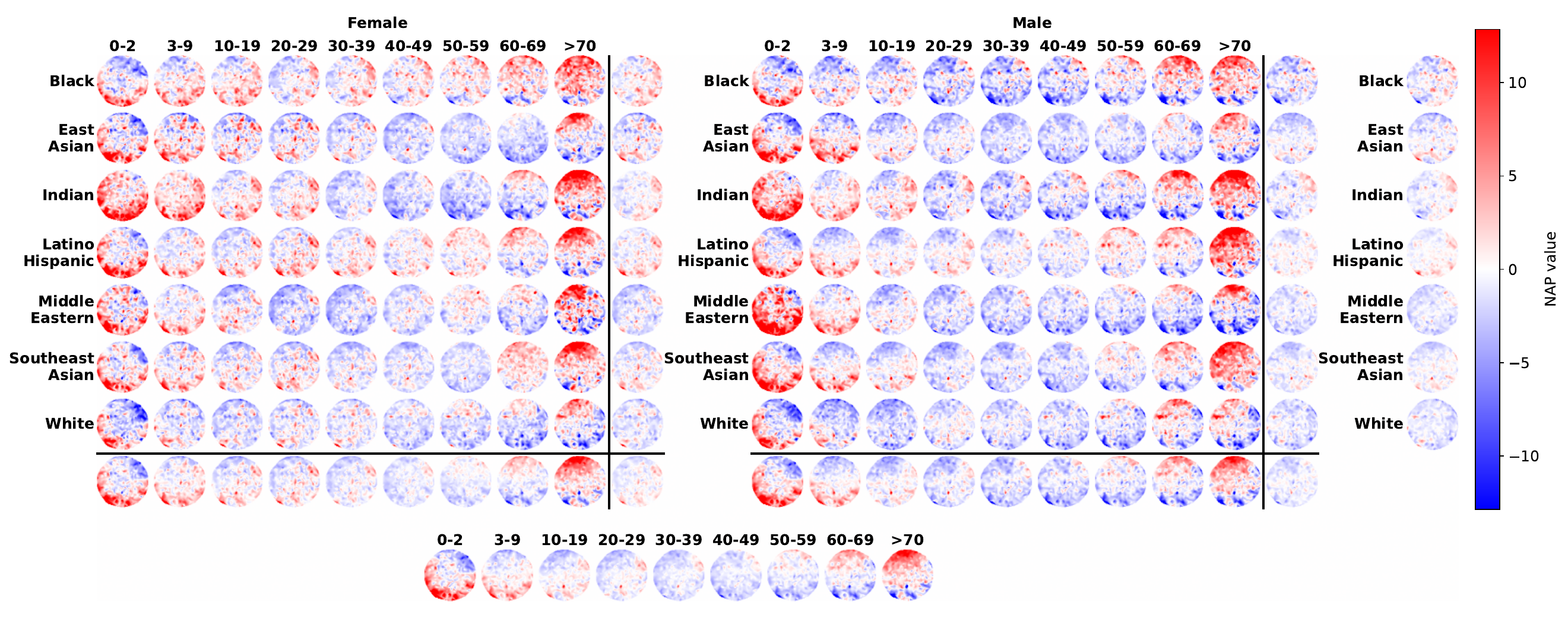}
	\caption{Topographic activation maps for all subgroups in VGG16 layer 4 (``block4\_pool'').}
	\label{fig:topo_VGG16_layer004}
\end{figure}

We can validate our findings by visualizing activations in \Cref{fig:topo_VGG16_layer004} and investigating their similarities.
Overall, groups show highest variation in the ``age'' dimension, where close age groups have similar activations.
There are differences between ``race'' groups, too, but not consistently across ages and genders.
For example, ``30--39, Male'' activations are visually very similar across races, but are better distinguishable for ``60--69, Male'.
Between genders, there is little difference, with exceptions like ``Indian, 60--69'' with higher activation for ``Male'' than for ``Female''.
This example also indicates, why age errors are mostly made for the ``Indian, Male`` groups because the similarity of ``Indian, 60--69'' and ``Indian, $>$70'' is clearly higher for ``Male'' than for ``Female''.
``Indian, $>$70, Male'' might be a common wrong classification because its activations are similar to several other groups.
However, there also are errors that cannot be explained by high activation similarity, like wrongly classifying ``White, $>$70, Male'' as ``Indian, $>$70, Male''.

\captionsetup[subfigure]{format=plain}
\begin{figure}[!b]
	\centering
	\begin{subfigure}{0.4\linewidth}
		\includegraphics[width=\linewidth]{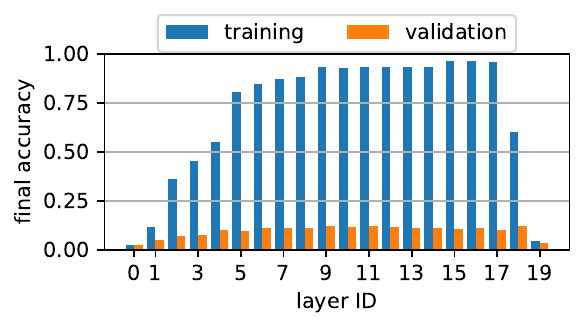}
		\caption{Accuracies using average pooling}
	\end{subfigure}
	\begin{subfigure}{0.4\linewidth}
		\includegraphics[width=\linewidth]{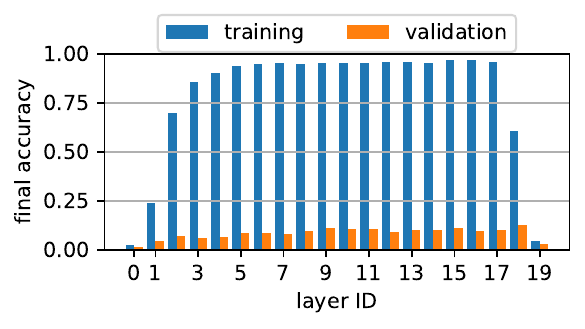}
		\caption{Accuracies using subsampling}
	\end{subfigure}
	
	\begin{subfigure}{0.24\linewidth}
		\includegraphics[width=0.9\linewidth]{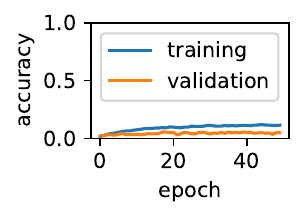}
		\caption{Avg. pooling, layer 1}
	\end{subfigure}
	\begin{subfigure}{0.24\linewidth}
		\includegraphics[width=0.9\linewidth]{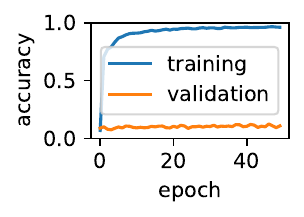}
		\caption{Avg. pooling, layer 15}
	\end{subfigure}
	\begin{subfigure}{0.24\linewidth}
		\includegraphics[width=0.9\linewidth]{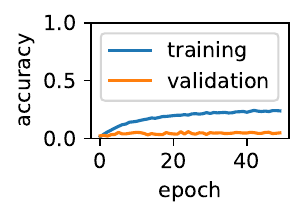}
		\caption{Subsampling, layer 1}
	\end{subfigure}
	\begin{subfigure}{0.24\linewidth}
		\includegraphics[width=0.9\linewidth]{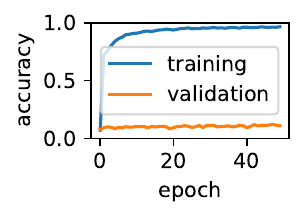}
		\caption{Subsampling, layer 15}
	\end{subfigure}
	\caption{Classifier probe accuracies along the ResNet50 layers and selected learning curves with different downsampling approaches.}
	\label{fig:ResNet50_probe_overview}
	\label[pluralfig]{figs:ResNet50_probe_overview}
\end{figure}

\subsection{ResNet50}

\subsubsection{Layer Overview}
Accuracies at the end of the training time along the layers are shown in \Cref{figs:ResNet50_probe_overview}a--b.
We observe that the linear separability increases until layer 17 and decreases toward the output layer again.
In earlier layers, the classifiers perform worse with average pooled feature maps than with subsampled ones.
This indicates that there is more information lost by applying pooling than by subsampling, particularly for large feature maps in earlier layers.
Using subsampling, groups can be classified from representations in layers 6--17 with a very high training accuracy but a low validation accuracy.
Performances in this layer range are very similar, therefore we choose layer 15 as a representative layer with high relative training and validation accuracy of the classifier.

High training accuracy with low performance on the validation accuracy indicates overfitting.
By inspecting the learning curves we can refute that this is due to long training time, as shown for exemplary layers in \Cref{figs:ResNet50_probe_overview}c--f.
In neither of the layers and downsampling strategies, the validation accuracy experiences a tipping point.

\subsubsection{Classification Errors}
For the fifteenth layer, we investigate the most frequent classification errors, shown in \Cref{tbl:errors_ResNet50_layer015}.
Notably, none of the frequent errors is based on a wrong gender and errors are wrong in both race and age.
We observe age errors across different age groups.
They are almost always wrongly classified as the adjacent age, like classifying ``40--49'' as either ``30--39'' or ``50--59''. 
Such errors are understandable because, although faces change with age, they do not strictly indicate an age, particularly at the boundaries between two age ranges.
Most errors for the ``race'' variable are made for ``Latino\_Hispanic'' groups which are wrongly predicted as ``White'', ``Indian'' or ``Middle Eastern''.

\begin{table}[!hbt]
	\def\arraystretch{1.1}
	\centering
		\begin{tabular}{cccllcccllr}
			\toprule
			\multicolumn{3}{c}{label} &&& \multicolumn{3}{c}{prediction} &&& error rate for  \\
			race & age & gender &&& race & age & gender &&& target label \\
			\midrule
			White & \textbf{30--39} & Male &&& White & \textbf{20--29} & Male &&& 4.35\% \\
			\textbf{Latino\_Hispanic} & \textbf{20--29} & Female &&& \textbf{Indian} & \textbf{10--19} & Female &&& 3.48\% \\
			\textbf{Middle Eastern} & \textbf{30--39} & Male &&& \textbf{White} & \textbf{20--29} & Male &&& 3.48\% \\
			\textbf{Latino\_Hispanic} & 3--9 & Female &&& \textbf{Middle Eastern} & 3--9 & Female &&& 3.48\% \\
			Indian & \textbf{40--49} & Female &&& Indian & \textbf{50--59} & Female &&& 3.48\% \\
			\textbf{Latino\_Hispanic} & 40--49 & Male &&& \textbf{White} & 40--49 & Male &&& 3.48\% \\
			\textbf{Latino\_Hispanic} & \textbf{40--49} & Male &&& \textbf{White} & \textbf{20--29} & Male &&& 3.48\% \\
			\textbf{Latino\_Hispanic} & \textbf{40--49} & Male &&& \textbf{Indian} & \textbf{30--39} & Male &&& 3.48\% \\
			\textbf{Middle Eastern} & \textbf{50--59} & Female &&& \textbf{White} & \textbf{60--69} & Female &&& 3.48\% \\
			\textbf{Southeast Asian} & 3--9 & Female &&& \textbf{East Asian} & 3--9 & Female &&& 2.61\% \\
			\bottomrule
		\end{tabular}
\vspace{0.2cm}
	\caption{Frequent training set errors of a linear classifier probe trained on layer 15 of ResNet50. Categories in bold print indicate differences between annotation and predicted group.}
	\label{tbl:errors_ResNet50_layer015}
\end{table}

\vspace{-1cm}

\subsubsection{Activation Visualization}
\begin{figure}[!htb]
	\centering
	\includegraphics[width=\linewidth]{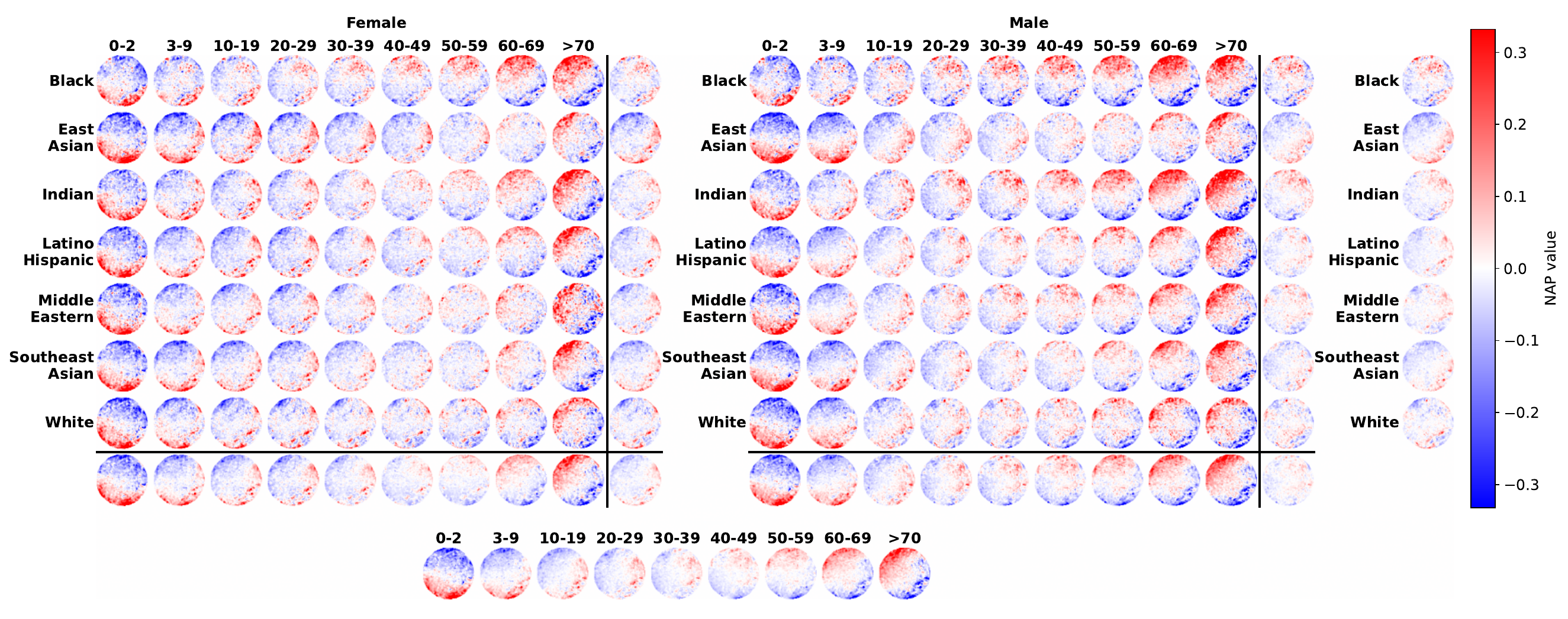}
	\caption{Topographic activation maps for all subgroups in ResNet50 layer 15 (``conv5\_block1\_out'').}
	\label{fig:topo_ResNet50_layer015}
\end{figure}

To support our findings, we show a visualization of ResNet50 activations in layer 15 in \Cref{fig:topo_ResNet50_layer015}.
The observed confusion of age groups fits to the patterns in the visualization where activations of groups of adjacent ages are similar.
In addition, there appears to be a stronger difference between ages ``50--59'' to ``60--69''.
Comparing races, we do not find such clear differences, which supports the potential to confuse race categories.
However, we do not find a clear reason why the classifier particularly confuses ``Latino\_Hispanic'' groups as often.
Finally, the lack of errors in the gender variable is reasonable because most groups can be easily distinguished between the ``Female'' and ``Male'' category, particularly for the middle ages.

\vspace{-0.5cm}

\captionsetup[subfigure]{format=plain}
\begin{figure}[!b]
	\centering
	\begin{subfigure}{0.4\linewidth}
		\includegraphics[width=\linewidth]{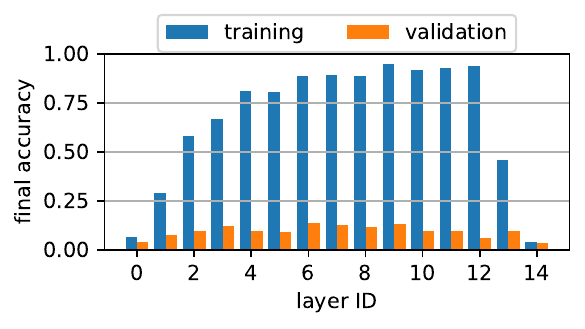}
		\caption{Accuracies using average pooling}
	\end{subfigure}
	\begin{subfigure}{0.4\linewidth}
		\includegraphics[width=\linewidth]{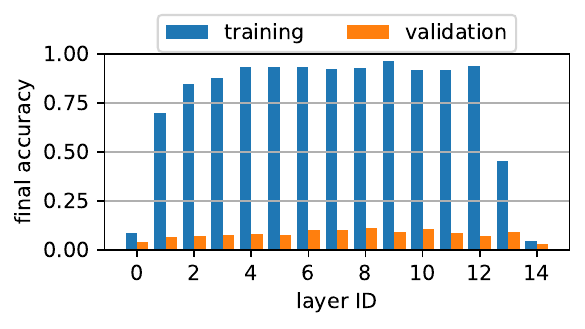}
		\caption{Accuracies using subsampling}
	\end{subfigure}
	
	\begin{subfigure}{0.24\linewidth}
		\includegraphics[width=0.9\linewidth]{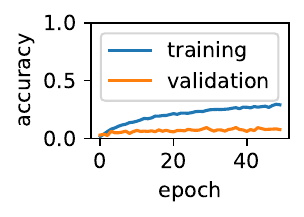}
		\caption{Avg. pooling, layer 1}
	\end{subfigure}
	\begin{subfigure}{0.24\linewidth}
		\includegraphics[width=0.9\linewidth]{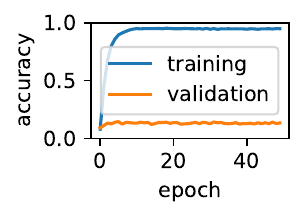}
		\caption{Avg. pooling, layer 9}
	\end{subfigure}
	\begin{subfigure}{0.24\linewidth}
		\includegraphics[width=0.9\linewidth]{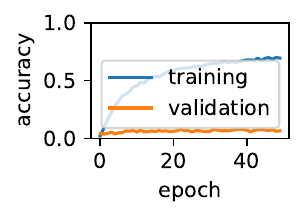}
		\caption{Subsampling, layer 1}
	\end{subfigure}
	\begin{subfigure}{0.24\linewidth}
		\includegraphics[width=0.9\linewidth]{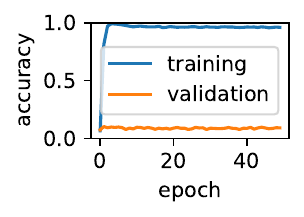}
		\caption{Subsampling, layer 9}
	\end{subfigure}
	
	\caption{Classifier probe accuracies along the InceptionV3 layers and selected learning curves with different downsampling approaches.}
	
	\label{fig:InceptionV3_probe_overview}
	\label[pluralfig]{figs:InceptionV3_probe_overview}
	
\end{figure}

\subsection{InceptionV3}

\subsubsection{Layer Overview}
Linear classifier accuracies for InceptionV3 are shown in \Cref{figs:InceptionV3_probe_overview}a--b.
The downsampling strategy only differs in lower layers, where average pooling leads to worse separability than subsampling.
The classifier probes show a very high training and low validation accuracy in all layers except the input and the two final layers.
In InceptionV3, classifiers perform best on the 9th layer, for which we will investigate classification errors.
Classifier probe performances do not show local validation accuracy maxima, as exemplarily shown in \Cref{figs:InceptionV3_probe_overview}c--f.

\subsubsection{Classification Errors}

\Cref{tbl:errors_InceptionV3_layer009} shows the most frequent classification errors for the ninth layer of InceptionV3.
The errors are mainly made for the age but misclassifications of the race is common, as well.
Different target age groups from 30 and older are affected by these errors, which are wrongly classified as groups with ages ``40--69''.
Mostly, the age group is wrong by predicting the next older or younger group.
However, some groups of ``$>$70'' are misclassified further apart as ages ``50--59'' or ``40--49''.
Errors made in regarding ``race'' are in most cases missclassfications as ``Latino\_Hispanic''.
A particularly common error type is the classfication as ``Latino\_Hispanic, 40--49, Male'' for 5 of 10 error types.

\begin{table}[!htb]
	\def\arraystretch{1.1}
		\centering
		\begin{tabular}{cccllcccllr}
			\toprule
			\multicolumn{3}{c}{label} &&& \multicolumn{3}{c}{prediction} &&& error rate for  \\
			race & age & gender &&& race & age & gender &&& target label \\
			\midrule
			Latino\_Hispanic & \textbf{30--39} & Male &&& Latino\_Hispanic & \textbf{40--49} & Male &&& 6.09\% \\
			East Asian & \textbf{$>$70} & Male &&& East Asian & \textbf{60--69} & Male &&& 5.13\% \\
			\textbf{Middle Eastern} & \textbf{50--59} & Male &&& \textbf{Latino\_Hispanic} & \textbf{40--49} & Male &&& 4.35\% \\
			Middle Eastern & 0--2 & \textbf{Female} &&& Middle Eastern & 0--2 & \textbf{Male} &&& 4.00\% \\
			Black & \textbf{$>$70} & Male &&& Black & \textbf{50--59} & Male &&& 4.00\% \\
			\textbf{Black} & \textbf{30--39} & Male &&& \textbf{Latino\_Hispanic} & \textbf{40--49} & Male &&& 3.48\% \\
			\textbf{Southeast Asian} & 0--2 & Female &&& \textbf{East Asian} & 0--2 & Female &&& 2.94\% \\
			Latino\_Hispanic & \textbf{$>$70} & \textbf{Female} &&& Latino\_Hispanic & \textbf{40--49} & \textbf{Male} &&& 2.04\% \\
			\textbf{Indian} & \textbf{$>$70} & Male &&& \textbf{Latino\_Hispanic} & \textbf{40--49} & Male &&& 1.85\% \\
			\textbf{White} & 0--2 & Female &&& \textbf{Latino\_Hispanic} & 0--2 & Female &&& 1.74\% \\
			\bottomrule
		\end{tabular}
	\vspace{0.2cm}
	\caption{Frequent training set errors of a linear classifier probe trained on layer 9 of InceptionV3. Categories in bold print indicate differences between annotation and predicted group.}
	\label{tbl:errors_InceptionV3_layer009}
\end{table}

\vspace{-1cm}

\subsubsection{Activation Visualization}
\begin{figure}[!htb]
	\centering
	\includegraphics[width=\linewidth]{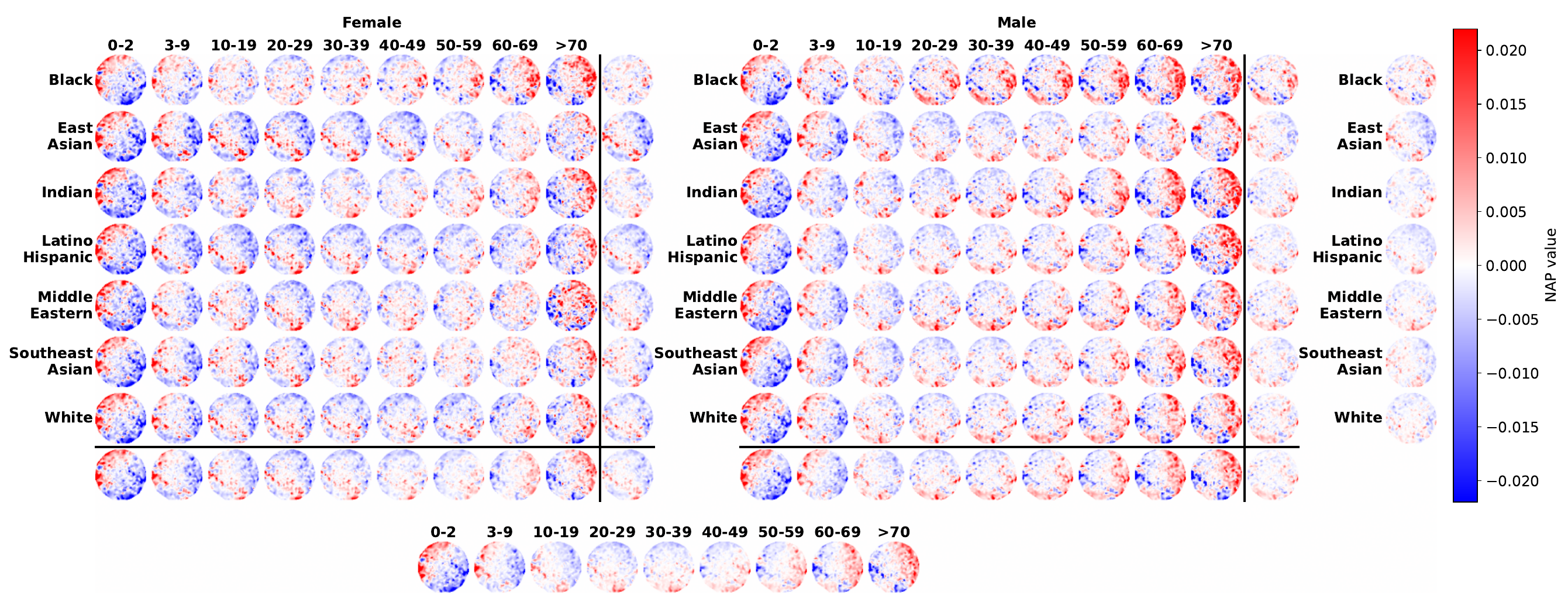}
	\caption{Topographic activation maps for all subgroups in InceptionV3 layer 9 (``mixed7'').}
	\label{fig:topo_InceptionV3_layer009}
\end{figure}

In \Cref{fig:topo_InceptionV3_layer009}, we visualize activations in the ninth layer of InceptionV3.
Frequently, groups are wrongly classified as age ``40--49''.
From inspecting the activations, we find that, for the ``Female'' category, this age group is similar to younger ages down to 20 years for most races.
In the ``Male'' category, age group ``40--49'' is similar to a wider range of ages from 20--59.
Especially the similarities in the ``Male'' category explain the common age misclassification because ``40--49'' might be an average activation of these similar groups.
Generally, there is a continuous change of activations along the age groups such that adjacent ages are similar to each other.
This effect is reflected in the errors that mostly are only wrong by one age group.
However, the youngest as well as the oldest age groups (``0--2'' and ``$>$70'') are often better distinguishable from their closest age neighbor than other age categories.

Several groups are frequently misclassified as ``Latino\_Hispanic'', which indicates that this race category is similar to others.
We can confirm this from the visualization although it is not clear why exactly this group is the common prediction.
Our hypothesis is that ``Latino\_Hispanic'' is approximately the average of several similar groups and leads to confusion of categories.
More specifically, a common error is the classification as ``Latino\_Hispanic, 40--49, Male''.
Activations for this group are similar to the other gender, to other ages from 20--59 and to races ``Middle Eastern'' and ``White''.
This agrees with most groups that are frequently classified as ``Latino\_Hispanic, 40--49, Male'', but not all.

\subsection{Summary and Model Comparison}
\label{sec:result_summary}

Each model shows similar classifier performance along the layers.
Input and output layer activations are not sufficient to linearly separate the intersectional groups.
Layers close to input or output lead to medium classification performance.
For all central layers, the groups are well linearly separable.
These results suggest that biases are encoded in the representations, however, not general enough for a linear classifier to learn a reliable separation.
In practice, classifiers are commonly multi-layer and non-linear and would be more likely to learn generalized patterns from the activations.

From topographic map visualization, we observed that activations along the age groups develop continuously.
This is reasonable because age is only weakly connected to the appearance of a person and therefore, a grouping by age necessarily leads to overlapping groups.
Moreover, we found that activation differences between genders mostly emerge for middle ages from 10--59.
We consider this a plausible observation, too, because gendered visual cues like hair styles, make-up or facial hair are more prevalent in these age groups.

We found that, across models, the age difference is always most pronounced compared to race and gender, particularly confirmed by the activation visualization.
We suspect that age differences in facial images are primarily driven by change of skin texture with age, like wrinkles, or facial shape properties, like round faces with large eyes for infants.
The hypothesis that such features are encoded in the models' representations is in line with previous research on general object detection which indicates that \acp{DNN} focus on structured patterns to perform predictions \cite{baker2018deep}.

\section{Conclusion}

In this work, we investigated how different pre-trained ImageNet classifiers activate for facial images comparing the sensitive variables ``race'', ``age'' and ``gender'' in an intersectional approach.
We proposed to analyze the representational similarity under the aspect of intersectional biases with linear classifier probes and activation visualization.

Linear classifier probes could distinguish training data, but did not generalize to unseen data. 
A non-linear classifier, however, would likely be able to facilitate encoded biases from almost any layer.
Therefore, we argue that linear classifier probes are useful but using them as the only measure is insufficient to analyze biases in representations.

The investigated models showed many commonalities in the activation similarities of the groups and consequently in the potential biases.
In particular, we found that differences are high between age categories, especially for the youngest and oldest ages in our data, the groups ``0--2'' and ``$>$70''.
Further, the models tended to confuse certain ethnicities due to activation similarity, mostly between the categories ``Latino\_Hispanic'', ``White'', ``Indian'' and ``Middle Eastern''.

In summary, our experiments indicate that pre-trained ImageNet classifiers, regardless of the architecture, strongly encode age information in their representations. 
This is in agreement with a recent related study that follows a clustering-based analysis \cite{krug2025intersectional}.
Racial biases are present, too, but are less pronounced. 
The models tend to confuse certain ethnicities while distinguishing the others.
Gender information is encoded in the more complex models and only for middle ages.
Note that all results in this study are specific to facial images and their features.

\subsubsection{Future Work} 
To investigate whether the observed effects arise from ImageNet specifically, we aim to analyze models trained on different data sets.
Similarly, the results could vary for other evaluation data than FairFace which we will examine in future.
In addition to \acp{CNN} which we focused on in this work, we will, for example, investigate Vision Transformers \cite{dosovitskiy2020image} and models trained with biologically more plausible learning schemes \cite{StrickerRoehrbeinKnoblauch} to further isolate the influence of the type of architecture.
Moreover, we plan to enrich the methodology of our analyses.
For example, a bias for one intersectional dimension might overshadow others which poses the risk to not identify some less pronounced biases.
This implies that future studies could also investigate the dimensions separately, like analyzing racial bias for every age group and gender individually.

%
%
%
%
 \bibliographystyle{abbrv}
 \bibliography{sample-base}

\begin{thebibliography}{10}

\bibitem{ahn-oh-2021-mitigating}
J.~Ahn and A.~Oh.
\newblock Mitigating language-dependent ethnic bias in {BERT}.
\newblock In {\em Proceedings of the 2021 Conference on Empirical Methods in
  Natural Language Processing (EMNLP)}, pages 533--549, Online and Punta Cana,
  Dominican Republic, Nov. 2021. Association for Computational Linguistics.

\bibitem{Alain2017}
G.~Alain and Y.~Bengio.
\newblock Understanding intermediate layers using linear classifier probes.
\newblock In {\em International Conference on Learning Representations (ICLR),
  Workshop Track Proceedings}, 2017.

\bibitem{baker2018deep}
N.~Baker, H.~Lu, G.~Erlikhman, and P.~J. Kellman.
\newblock Deep convolutional networks do not classify based on global object
  shape.
\newblock {\em PLoS computational biology}, 14(12):e1006613, 2018.

\bibitem{bolukbasi2016man}
T.~Bolukbasi, K.-W. Chang, J.~Y. Zou, V.~Saligrama, and A.~T. Kalai.
\newblock {Man is to computer programmer as woman is to homemaker? Debiasing
  Word Embeddings}.
\newblock In {\em Advances in Neural Information Processing Systems},
  volume~29, pages 4349--4357, 2016.

\bibitem{pmlr-v81-buolamwini18a}
J.~Buolamwini and T.~Gebru.
\newblock {Gender Shades: Intersectional Accuracy Disparities in Commercial
  Gender Classification}.
\newblock In {\em Proceedings of the 1st Conference on Fairness, Accountability
  and Transparency}, volume~81, pages 77--91. PMLR, 23--24 Feb 2018.

\bibitem{carter2019activation}
S.~Carter, Z.~Armstrong, L.~Schubert, I.~Johnson, and C.~Olah.
\newblock Activation atlas.
\newblock {\em Distill}, 4(3):e15, 2019.

\bibitem{deng2009imagenet}
J.~Deng, W.~Dong, R.~Socher, L.-J. Li, K.~Li, and L.~Fei-Fei.
\newblock {Imagenet: A large-scale hierarchical image database}.
\newblock In {\em 2009 IEEE Conference on Computer Vision and Pattern
  Recognition (CVPR)}, pages 248--255. Ieee, 2009.

\bibitem{devlin2019bert}
J.~Devlin, M.-W. Chang, K.~Lee, and K.~Toutanova.
\newblock {BERT: Pre-training of Deep Bidirectional Transformers for Language
  Understanding}.
\newblock {\em arXiv preprint arXiv:1810.04805}, 2019.

\bibitem{dosovitskiy2020image}
A.~Dosovitskiy, L.~Beyer, A.~Kolesnikov, D.~Weissenborn, X.~Zhai,
  T.~Unterthiner, M.~Dehghani, M.~Minderer, G.~Heigold, S.~Gelly, et~al.
\newblock An image is worth 16x16 words: Transformers for image recognition at
  scale.
\newblock {\em arXiv preprint arXiv:2010.11929}, 2020.

\bibitem{Erhan2009}
D.~Erhan, Y.~Bengio, A.~Courville, and P.~Vincent.
\newblock Visualizing higher-layer features of a deep network.
\newblock {\em University of Montreal}, 1341(3):1, 2009.

\bibitem{Fiacco2019}
J.~Fiacco, S.~Choudhary, and C.~Rose.
\newblock Deep neural model inspection and comparison via functional neuron
  pathways.
\newblock In {\em Annual Meeting of the Association for Computational
  Linguistics (ACL)}, pages 5754--5764, 2019.

\bibitem{goodfellow2016deep}
I.~Goodfellow, Y.~Bengio, A.~Courville, and Y.~Bengio.
\newblock {\em Deep learning}, volume~1.
\newblock MIT press Cambridge, 2016.

\bibitem{he2016deep}
K.~He, X.~Zhang, S.~Ren, and J.~Sun.
\newblock Deep residual learning for image recognition.
\newblock In {\em Proceedings of the IEEE conference on computer vision and
  pattern recognition}, pages 770--778, 2016.

\bibitem{hohman2019s}
F.~Hohman, H.~Park, C.~Robinson, and D.~H.~P. Chau.
\newblock Summit: Scaling deep learning interpretability by visualizing
  activation and attribution summarizations.
\newblock {\em IEEE Transactions on Visualization and Computer Graphics},
  26(1):1096--1106, 2019.

\bibitem{hort2023bias}
M.~Hort, Z.~Chen, J.~M. Zhang, M.~Harman, and F.~Sarro.
\newblock Bias mitigation for machine learning classifiers: A comprehensive
  survey.
\newblock {\em ACM Journal on Responsible Computing}, 2023.

\bibitem{karkkainenfairface}
K.~Karkkainen and J.~Joo.
\newblock {FairFace: Face Attribute Dataset for Balanced Race, Gender, and Age
  for Bias Measurement and Mitigation}.
\newblock In {\em Proceedings of the IEEE/CVF Winter Conference on Applications
  of Computer Vision}, pages 1548--1558, 2021.

\bibitem{Kim2018}
B.~Kim, M.~Wattenberg, J.~Gilmer, C.~Cai, J.~Wexler, F.~Viegas, et~al.
\newblock {Interpretability Beyond Feature Attribution: Quantitative Testing
  with Concept Activation Vectors (TCAV)}.
\newblock In {\em Proceedings of the International Conference on Machine
  Learning (ICML)}, pages 2668--2677, 2018.

\bibitem{kim2022squeezeformer}
S.~Kim, A.~Gholami, A.~Shaw, N.~Lee, K.~Mangalam, J.~Malik, M.~W. Mahoney, and
  K.~Keutzer.
\newblock Squeezeformer: An efficient transformer for automatic speech
  recognition.
\newblock In {\em Advances in Neural Information Processing Systems},
  volume~35, pages 9361--9373, 2022.

\bibitem{Kindermans2018}
P.-J. Kindermans, K.~T. Schütt, M.~Alber, K.-R. Müller, D.~Erhan, B.~Kim, and
  S.~Dähne.
\newblock Learning how to explain neural networks: {PatternNet and
  PatternAttribution}.
\newblock In {\em International Conference on Learning Representations (ICLR)},
  2018.

\bibitem{krug2024thesis}
V.~Krug.
\newblock {Neuroscience-Inspired Analysis and Visualization of Deep Neural
  Networks}, 2024.

\bibitem{krug2023topomapsbias}
V.~Krug, C.~Olson, and S.~Stober.
\newblock {Visualizing Bias in Activations of Deep Neural Networks as
  Topographic Maps}.
\newblock In {\em Proceedings of the 1st Workshop on Fairness and Bias in AI
  (AEQUITAS 2023) co-located with 26th European Conference on Artificial
  Intelligence (ECAI 2023) Kraków, Poland}. CEUR-WS, 2023.

\bibitem{krug2023visualizing}
V.~Krug, R.~K. Ratul, C.~Olson, and S.~Stober.
\newblock {Visualizing Deep Neural Networks with Topographic Activation Maps}.
\newblock In {\em HHAI 2023: Augmenting Human Intellect}, pages 138--152. IOS
  Press, 2023.

\bibitem{krug2025intersectional}
V.~Krug, F.~Röhrbein, and S.~Stober.
\newblock {Intersectional Bias Quantification in Facial Image Processing with
  Pre-Trained ImageNet Classifiers}.
\newblock In {\em 2025 International Joint Conference on Neural Networks
  (IJCNN)}, pages 1--8, 2025.

\bibitem{li2021detecting}
B.~Li, H.~Peng, R.~Sainju, J.~Yang, L.~Yang, Y.~Liang, W.~Jiang, B.~Wang,
  H.~Liu, and C.~Ding.
\newblock {Detecting Gender Bias in Transformer-based Models: A Case Study on
  BERT}.
\newblock {\em arXiv preprint arXiv:2110.15733}, 2021.

\bibitem{mcinnes2020umap}
L.~McInnes, J.~Healy, and J.~Melville.
\newblock {UMAP: Uniform Manifold Approximation and Projection for Dimension
  Reduction}.
\newblock {\em arXiv preprint arXiv:1802.03426}, 2020.

\bibitem{mcinnes2018umap-software}
L.~McInnes, J.~Healy, N.~Saul, and L.~Grossberger.
\newblock {UMAP: Uniform Manifold Approximation and Projection}.
\newblock {\em The Journal of Open Source Software}, 3(29):861, 2018.

\bibitem{Morcos2018a}
A.~S. Morcos, M.~Raghu, and S.~Bengio.
\newblock Insights on representational similarity in neural networks with
  canonical correlation.
\newblock {\em arXiv preprint arXiv:1806.05759}, 2018.

\bibitem{Mordvintsev2015}
A.~Mordvintsev, C.~Olah, and M.~Tyka.
\newblock Inceptionism: Going deeper into neural networks.
\newblock {\em Google Research Blog. Retrieved June}, 20(14):5, 2015.

\bibitem{Nagamine2015}
T.~Nagamine, M.~L. Seltzer, and N.~Mesgarani.
\newblock Exploring how deep neural networks form phonemic categories.
\newblock In {\em Conference of the International Speech Communication
  Association (Interspeech)}, 2015.

\bibitem{Osindero2008}
S.~Osindero and G.~E. Hinton.
\newblock {Modeling image patches with a directed hierarchy of Markov random
  fields}.
\newblock In {\em Advances in Neural Information Processing Systems}, pages
  1121--1128, 2008.

\bibitem{park2021neurocartography}
H.~Park, N.~Das, R.~Duggal, A.~P. Wright, O.~Shaikh, F.~Hohman, and D.~H.~P.
  Chau.
\newblock Neurocartography: Scalable automatic visual summarization of concepts
  in deep neural networks.
\newblock {\em IEEE Transactions on Visualization and Computer Graphics},
  28(1):813--823, 2021.

\bibitem{salman2022does}
H.~Salman, S.~Jain, A.~Ilyas, L.~Engstrom, E.~Wong, and A.~Madry.
\newblock When does bias transfer in transfer learning?
\newblock {\em arXiv preprint arXiv:2207.02842}, 2022.

\bibitem{Selvaraju2017}
R.~R. Selvaraju, M.~Cogswell, A.~Das, R.~Vedantam, D.~Parikh, and D.~Batra.
\newblock {Grad-CAM: Visual Explanations From Deep Networks via Gradient-Based
  Localization}.
\newblock In {\em IEEE International Conference on Computer Vision (ICCV)},
  pages 618--626, 2017.

\bibitem{DBLP:journals/corr/SimonyanZ14a}
K.~Simonyan and A.~Zisserman.
\newblock Very deep convolutional networks for large-scale image recognition.
\newblock {\em arXiv preprint arXiv:1409.1556}, 2014.

\bibitem{StrickerRoehrbeinKnoblauch}
P.~Stricker, F.~Röhrbein, and A.~Knoblauch.
\newblock {Weight Perturbation and Competitive Hebbian Plasticity for Training
  Sparse Excitatory Neural Networks}.
\newblock In {\em 2024 International Joint Conference on Neural Networks
  (IJCNN)}, pages 1--8, 2024.

\bibitem{sweeney2013discrimination}
L.~Sweeney.
\newblock {Discrimination in Online Ad Delivery}.
\newblock {\em arXiv preprint arXiv:1301.6822}, 2013.

\bibitem{szegedy2016rethinking}
C.~Szegedy, V.~Vanhoucke, S.~Ioffe, J.~Shlens, and Z.~Wojna.
\newblock Rethinking the inception architecture for computer vision.
\newblock In {\em Proceedings of the IEEE conference on computer vision and
  pattern recognition}, pages 2818--2826, 2016.

\bibitem{torrey2010transfer}
L.~Torrey and J.~Shavlik.
\newblock Transfer learning.
\newblock In {\em Handbook of research on machine learning applications and
  trends: algorithms, methods, and techniques}, pages 242--264. IGI global,
  2010.

\bibitem{Yosinski2015}
J.~Yosinski, J.~Clune, A.~Nguyen, T.~Fuchs, and H.~Lipson.
\newblock {Understanding Neural Networks Through Deep Visualization}.
\newblock {\em arXiv preprint arXiv:1506.06579}, 2015.

\end{thebibliography}
%
%
%
%
%
\end{document}